\documentclass[10pt,A4paper]{article}

\usepackage{arxiv}
\usepackage{url}

\usepackage{tikz}
\usetikzlibrary{positioning}
\usetikzlibrary{decorations.pathmorphing,patterns}
\usetikzlibrary{calc,patterns,decorations.markings}
\usetikzlibrary{shapes,arrows}

\usepackage{pgfplots}
\usepackage{algorithm,algpseudocode}
\usepackage{algorithmicx}
\usepackage{authblk}
\usepackage{amsmath}
\usepackage{amsfonts}
\usepackage{gensymb}
\usepackage{siunitx}
\pgfplotsset{every axis plot/.style={thick}}

\title{Monitoring weeder robots and anticipating their functioning by using advanced topological data analysis}

\author[1]{Tarek Frahi}
\author[1,2,5]{Abel Sancarlos}
\author[3]{Matthieu Galle}
\author[3]{Xavier Beaulieu}
\author[2]{Anne Chambard}
\author[4]{Antonio Falc\'o}
\author[5]{El\'ias Cueto}
\author[1,2]{Francisco Chinesta}

\affil[1]{{\small ESI Group chair. PIMM Lab. ENSAM Institute of Technology. Paris, France.}}
\affil[2]{{\small ESI Group, 3bis rue Saarinen, 94528 Rungis CEDEX, France}}
\affil[3]{{\small VITIROVER, 6 lieu-dit, Simard, 33330 Saint-Emilion, France}}
\affil[4]{{\small ESI-CEU International Chair CEU-UCH, Departamento de Matematicas, Fisica y Ciencias Tecnologicas, Universidad Cardenal Herrera-CEU, San Bartolome 55, 46115 Alfara del Patriarca, Valencia, Spain}}
\affil[5]{{\small Aragon Institute of Engineering Research. Universidad de Zaragoza. Zaragoza, Spain.}}

\begin{document}

\maketitle

\begin{abstract}
The present paper aims at analyzing the topological content of the complex trajectories that weeder-autonomous robots follow in operation. We will prove that the topological descriptors of these trajectories are affected by the robot environment as well as by the robot state, with respect to maintenance operations. Topological Data Analysis will be used for extracting the trajectory descriptors, based on homology persistence. Then, appropriate metrics will be applied in order to compare that topological representation of the trajectories, for classifying them or for making efficient pattern recognition.

\end{abstract}

\section{Introduction}

Autonomous robots follow a number of rules introduced into their controllers \cite{R1,R2,R3}. However, when they interact with the environment, small variations may result in long-time unpredictable motion. This behavior is very usual in mechanics, characterizing systems exhibiting deterministic chaos.

\begin{figure}
    \centering
    \includegraphics[scale=0.65]{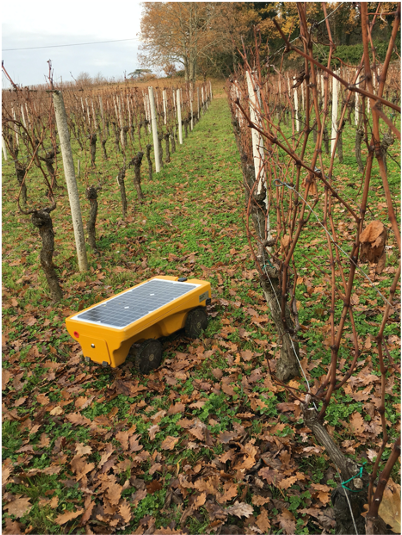}
    \caption{Weeder robot from VITIROVER {\em micro robotique viticole}}
    \label{robot}
\end{figure}

In the practical case addressed in the present paper, a weeder robot (usually a float of them) is expected to cover a patch of a vineyard, in an optimal manner. Here, ``optimal manner'' refers to the path-line that allows covering the whole patch in a minimum time. However, the ground orography has a significant variability, as well as the location of the grapes. Robots are aimed at colliding the grape foots in order to remove the grass around, and then numerous collisions following different directions are needed to ensure that all the grass around the grape foot is adequately removed. Figure \ref{robot} depicts one of these robots considered in the present study in operational conditions.

All the practical variability (ground, grape location, grass distribution and size, obstacles, ...) as well as the intrinsic sensibility of the dynamics to small variabilities in the physical and operational conditions, makes it impossible to define a deterministic robot trajectory. In these conditions, an almost random motion seems to be the most valuable alternative. 

In practice, to avoid under-performances characteristic of fully random motions, that random motion operating at the local scale is combined with a more global deterministic planning that tries to better control the vineyard coverage by sequencing the operation at the different local patches covering the whole domain.    

The present work does not aim at addressing such optimized operation conditions that will be addressed in a future publication under progress, but it aims at analyzing the data collected from a robot operating in different patches and under different conditions (with respect to the maintenance operations) in order to identify the existence of patterns able to identify the particular patch in which the robot operates, or to distinguish the different robot states with respect to the maintenance operations.

Having a sort of QR-code or identity card of each robot, when it operates within each patch, in a particular state (healthy or unhealthy), is of major relevance with respect to the predictive or operational maintenance of robots or floats of autonomous robots.

The present paper aims at analyzing the collected data in order to extract the maximum information that could serve for differentiating them, enabling unsupervised clustering and/or supervised classification, prior to any action concerning modeling using adapted regressions.

\section{Methods}

Using data clustering is almost straightforward, as soon as data is homogeneous and quantitatively expressible using integer or real numbers, enabling boolean or algebraic operations (addition, multiplication, ...) The interest of organizing data in groups, in a supervised or unsupervised manner, is that it is assumed that data belonging to a given group shares some qualities with the members of the group. 

When proceeding in an unsupervised manner, the only information to group the data consists of the distance among them. Data that remain close to each other are expected to share some properties or behavior. This is the rationale considered in the very popular {\it $k$-means} technique \cite{R4,R5}. However, the notion of proximity, leading to the derived concept of similarity, needs for the definition of a metric for comparison purposes. When data are well defined in a vector space, distances can be defined and data can be compared accordingly. In the case of supervised classification one is looking for the linear (or nonlinear) frontier separating the different groups on the basis of a quality or property that drives the data clustering. In this last case, the best frontier separating two groups of data is the one maximizing the distance of the available data to the frontier, in order to maximize the separation robustness. This is how support vector machine, SVM, works, for instance \cite{R6}.

In both cases (supervised and unsupervised) the existence of a metric enabling data comparison is assumed. However, very often data could be much more complex, as for example when it concerns heterogeneous information, possibly categorial or qualitative. This is for example the case when a manufactured part is described by its identity card consisting of the name of the employee involved in the operation, the designation of the employed materials (some of them given by its commercial name), the temperature of the oven in which the part was cured and the processing time. In that case, comparing two parts becomes quite controversial if the employed metric is not properly defined. In these circumstances, usually, metrics are learned from the existing training data, as is the case when using decision trees (or its random forest counterpart) \cite{R7,R8}, code-to-vector \cite{R9} or neural networks \cite{R10}.

The situation becomes even more extreme when data have a large and deep topology content. This is the case for example of time series or images of rich microstructures. These are usually encountered in material science when describing metamaterials (also called functional materials), or those exhibiting gradient of properties or mesoscopic architectures. Thus, even in nominal conditions,  time series will differ if they are compared from their respective values at each time instant. That is, two time series, even when they describe the same system in similar conditions, never match perfectly. Thus, they differ even if they resemble in a certain metric that should be learned. For example, our electrocardiogram measured during two consecutive minutes will exhibit a resemblance, but certainly both of them are not identical, thus making a perfect match impossible. A small variation will create a misalignment needing for metrics less sensible to these effects. The same rationale applies when comparing two profiles of a rough surface, two images of a foam taken in two close locations, ... they exhibit a resemblance even if they do not perfectly match.

Thus, techniques aiming at aligning data were proposed. In the case of time-series, Dynamic Time Warping, DTW \cite{R11,R12} has been successfully applied in many domains. The theory of optimal transport arose as a response to similar issues \cite{R13}.

Another route consists of renouncing to {\it align} the data, and focussing on extracting the adequate, goal-oriented descriptors of these complex data, enabling comparison, clustering, classification and modeling (from nonlinear regressions).

A first possibility consists of extracting the main statistical descriptors of time series or images (moments, correlations, covariograms, ...) \cite{R14}. Sometimes, data expressed in the usual space and time domains, are transformed into other spaces where their manipulation is expected to be simpler, like Fourier, Laplace, DCT, Wavelet, ... descriptions of data. The most valuable (in the sense given later) descriptions seem to be those maximizing sparsity. These are widely considered when using compressed sensing \cite{R15}, because it represents a compact, concise and complete way of representing data that seemed much more complex in the usual physical space (space and time).

The present work considers this last route, but uses a description based on the topology of data, described later, and successfully considered in our former works for addressing complex mesostructures \cite{R16}, time-series \cite{R17}, rough surfaces \cite{R18} and shapes \cite{R19}, with the aim of classifying and also constructing robust regressions expressing properties or performance from the input data expressed from its topological description.

The present study, when compared with our former developments, addresses a new and complex purpose: how the topology contained in the trajectory that an autonomous robot follows in a cloudy environment (where interactions limits the predictability horizon) can inform on the robot location (which patch into the whole vineyard) or the robot state (with respect to maintenance operations).

\subsection{Data description}

\begin{figure}
    \centering
    \includegraphics[scale=0.35]{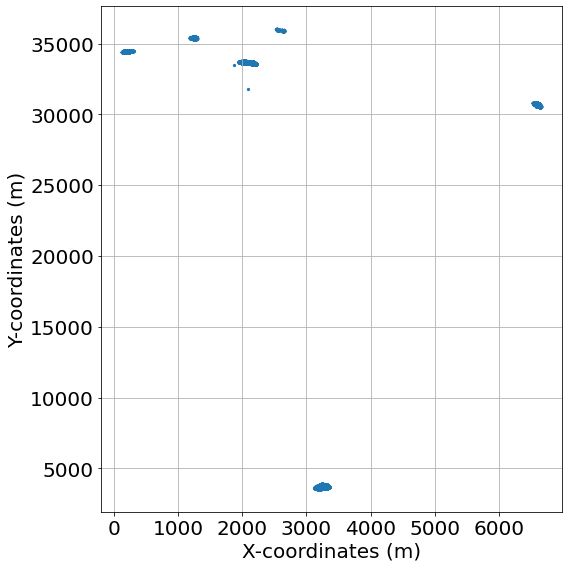}
    \caption{Location of the different patches}
    \label{fig:parcels}
\end{figure}

\begin{figure}
    \centering
    \includegraphics[scale=0.3]{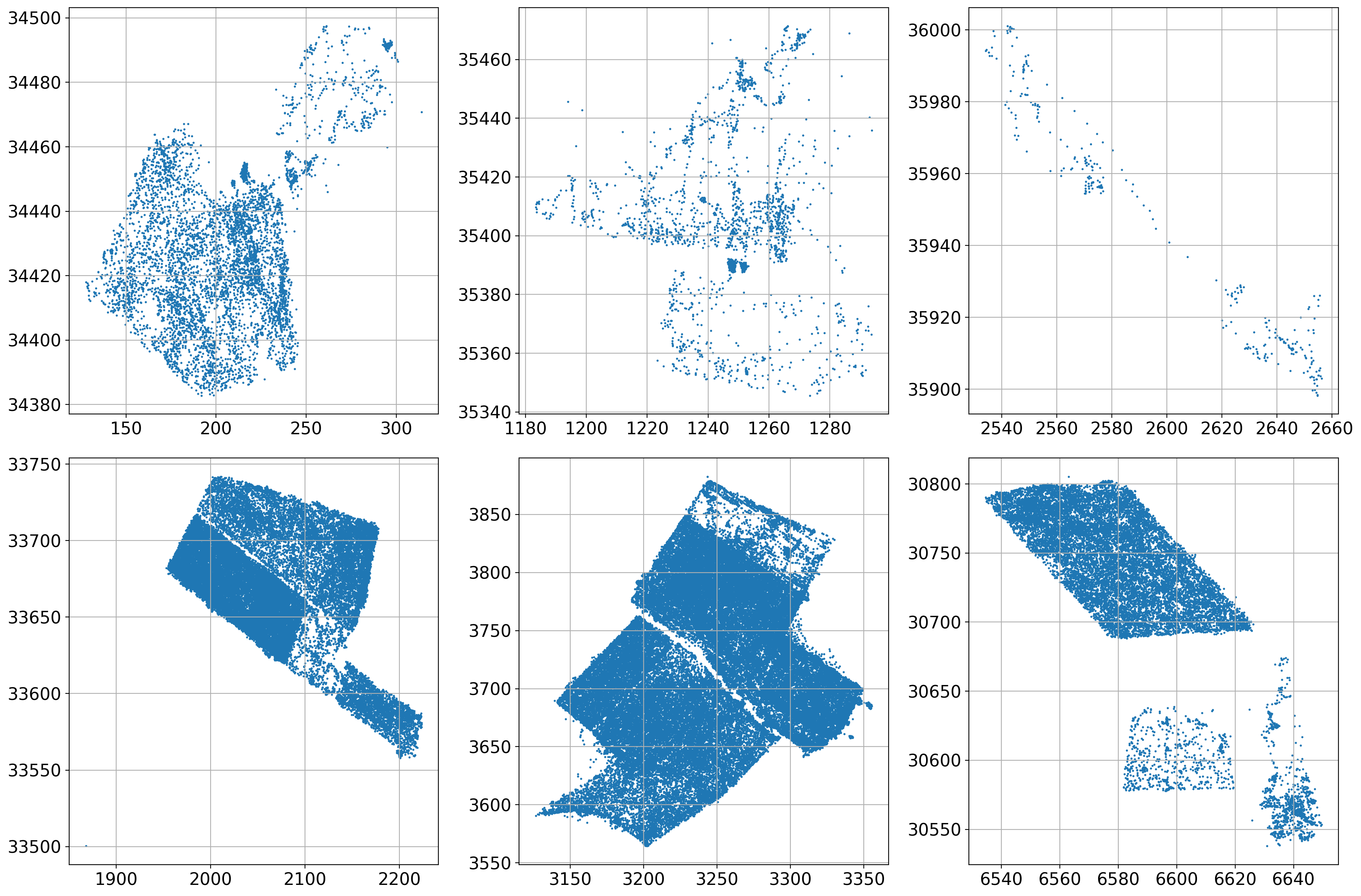}
    \caption{Robot trajectories in the six considered vineyard patches  (units in meters)}
    \label{fig:allparcels}
\end{figure}

In the study that follows, we consider a dataset consisting of the $x$ and $y$-coordinates, calculated from the GPS longitudes and latitudes, representing the recorded position of the robot at time $t$:
$$
\mathcal{D} = \{(x(t), y(t), t), \ t \in \mathcal{T}\}.
$$
These coordinates span six different disjoint geographical patches within the whole vineyard, as illustrated in Figure \ref{fig:parcels}, that have been recorded in a period of time $\mathcal{T}$ leading to the maps reported in Figure \ref{fig:allparcels} that reflects the robot's trajectory. 

Maintenance operations are also known and properly identified in the provided dataset. Thus, the dataset consists of a collection of $n$ discrete, finite and compact two-dimensional trajectories $\mathbb{S}_1,\dots,\mathbb{S}_n$.

\subsection{Geometrical Features}

\begin{figure}
    \centering
    \includegraphics[width=0.6\linewidth]{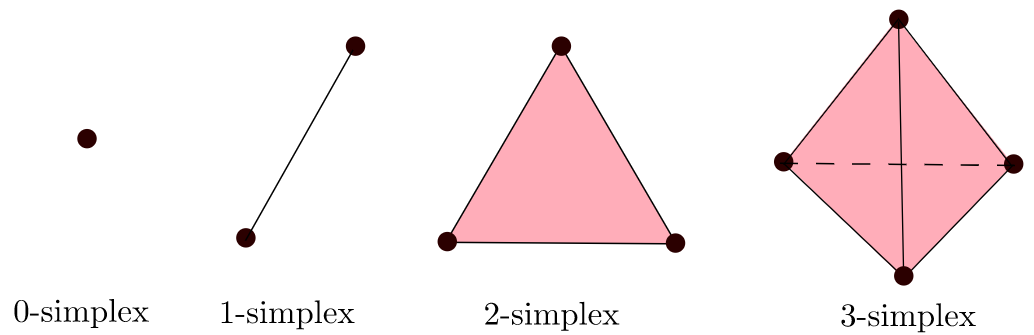}
    \caption{Simplices of different dimensions}
    \label{fig:simps}
\end{figure}

We are interested in extracting the geometrical and topological features of the trajectories in $\mathcal{D}$ across different scales. For that purpose, we introduce the so-called \emph{Rips filtration}. We construct a \emph{Rips complex} from simplices of varying dimensions that are generalizations of triangles of varying dimensions. More specifically, a $d$-simplex is the smallest convex set of $d+1$ points, $x_0,\ldots,x_{d}$ where $x_1-x_0,\ldots,x_{d}-x_0$ are linearly independent, as illustrated in Fig. \ref{fig:simps}. The so-called \emph{abstract simplicial complex} is a finite collection of sets that is closed under the subset relation, i.e., if $a \in A$ and $b \subset a$, then $b \in A$.

Let $\mathbb{S}$ be a trajectory, defined from a finite compact set of points in $\mathbb{R}^2$, and $\epsilon \geq 0$. The Rips complex of $\mathbb{S}$ at scale $\epsilon$, $\mathcal{R}_{\epsilon}(\mathbb{S})$, is the abstract simplicial complex consisting of all subsets of diameter up to $\epsilon$:
$$
\mathcal{R}_{\epsilon}(\mathbb{S}):= \{\sigma \subset \mathbb{S} \ | \ \text{diam}(\sigma) \leq \epsilon\},
$$
where the diameter of a set of points is the maximum distance between any two points in the set.

\begin{figure}[h!]
     \centering
         \includegraphics[scale=0.45]{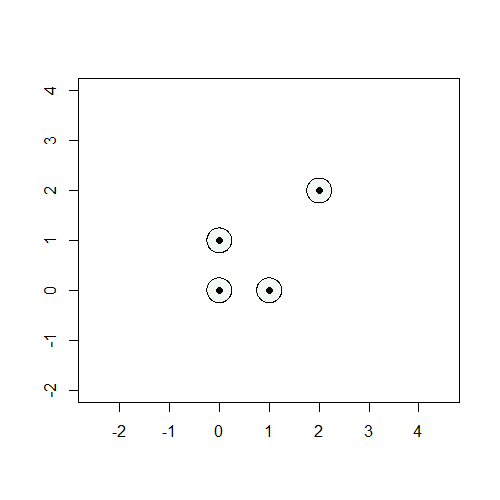}
         \includegraphics[scale=0.45]{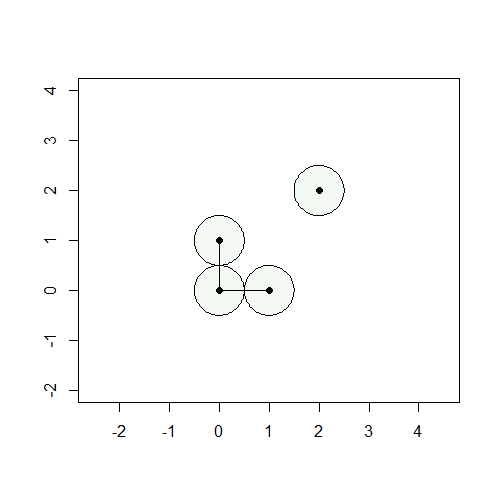}
         \includegraphics[scale=0.45]{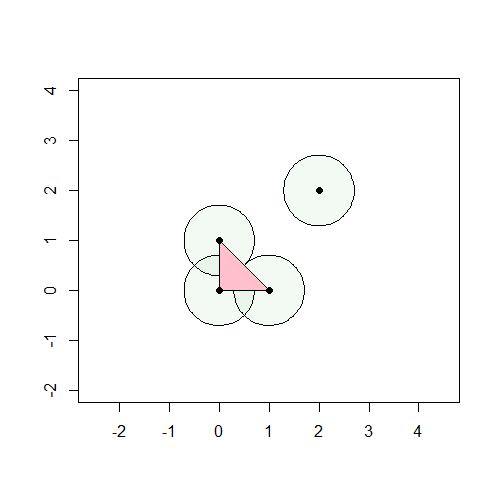}
         \includegraphics[scale=0.45]{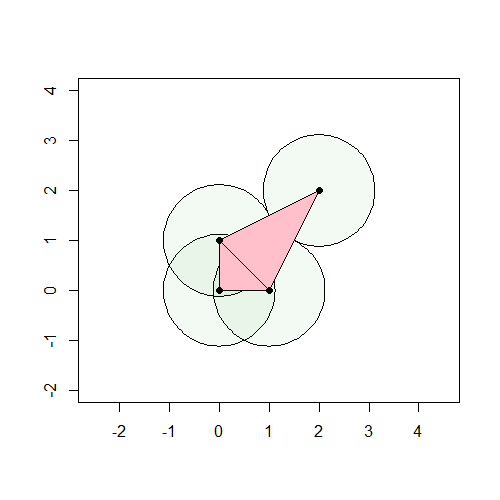}
        \caption{Example of Rips complex computation: (top-left) $\epsilon=0.5$; (top-right) $\epsilon=1$; (bottom-left) $\epsilon=1.4$; and (bottom-right) $\epsilon=2.3$.}
        \label{fig:rips}
\end{figure}

Geometrically, we can construct the Rips complex by considering balls of radius $\frac{\epsilon}{2}$, centered at each point in $\mathbb{S}$. Whenever $d$ balls have pairwise intersections, we add a $d-1$ dimensional simplex. An example of Rips complex is given in Fig. \ref{fig:rips}.


A \emph{filtration} of a simplicial complex $\mathcal{K}$ is a nested sequence of subcomplexes starting at the empty set and ending with the full simplicial complex
$$
\emptyset \subset \mathcal{K}_0 \subset \dots \subset \mathcal{K}.
$$
By varying the value of the scale parameter $\epsilon$, from $\epsilon_{\text{min}}=0$ to $\epsilon_{\text{max}} = \text{diam}(\mathbb{S})$
we get a family of nested  Rips complexes known as the Rips filtration.


\subsection{Persistent homology}

In order to have a more exhaustive view on how the features are changing across different scales, the appearance and disappearance of each feature within the filtration is tracked and coded into the homology groups $H_k(\mathbb{S})$, where $k$ is the homology dimension. The elements of a \emph{Homology Group} $H_k(\mathbb{S})$ are classes of chain of simplices (``packets'') in the Rips complex. The use of homology groups allows us to perform algebraic operations over the simplicial elements. The homology group $H_0(\mathbb{S})$ represents the vertices, while the homology group $H_1(\mathbb{S})$ represents the cycles (loops) formed in the simplicial complex. Since our data is in $\mathbb{R}^2$ we are only interested in $k=0$ and $k=1$.

Given a homology group, we can now define how to track the appearance of the features across different scales, by defining the homology group at a scale $\epsilon$, $H_k^{\epsilon}(\mathbb{S})$. It represents the classes of simplices as described previously, but taken from $\mathcal{R}_{\epsilon}(\mathbb{S})$. That is, the elements of  $\mathcal{R}_{\epsilon}(\mathbb{S})$ with a filtration value lower than $\epsilon$. This approach is known as the \emph{persistent homology}. It allows to quantify the appearance and disappearance of the features across the different scales (discretized by considering $m$ values related to $\epsilon_j$, $j=0, ..., m$) :
\begin{itemize}
\item For $H_0(\mathbb{S})$, the birth scale of all vertices is set to zero, while the death scale is the filtration value at which the vertex has been joined to another one by a segment.
\item For $H_1(\mathbb{S})$, the birth scale of a cycle is the filtration value at which a loop has been formed, while the death scale is the filtration value at which the interior of the loop has been covered.
\end{itemize}
We can formalize this as follows:
\begin{itemize}
\item The birth scale $b_{\gamma}$ of the feature $\gamma$ 
$$
b_{\gamma} = \min_{0 \leq j \leq m} \{\epsilon_j : \gamma \in H^{\epsilon_j}_k\}
$$
\item The death scale $d_{\gamma}$ of the feature $\gamma$ 
$$
d_{\gamma} = \max_{0 \leq j \leq m} \{\epsilon_j : \gamma \in H^{\epsilon_j}_k\}
$$
\end{itemize}
The persistence of the features throughout the scales can then be represented by the so-called \emph{persistence barcode} of $\mathbb{S}$. It is a histogram, where the bar associated to each feature starts at the birth scale and ends at the death scale.

\begin{figure}[h!]
	\centering
        \includegraphics[scale=0.45]{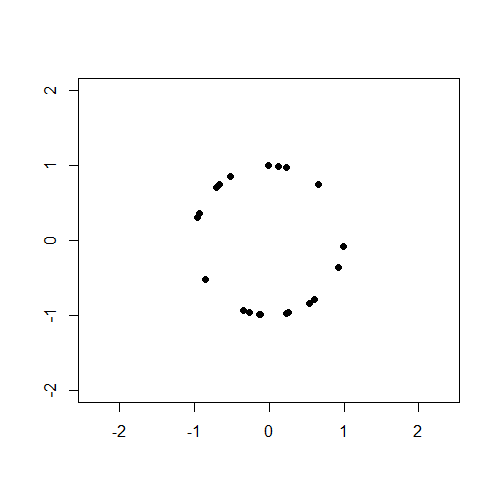}
         \includegraphics[scale=0.45]{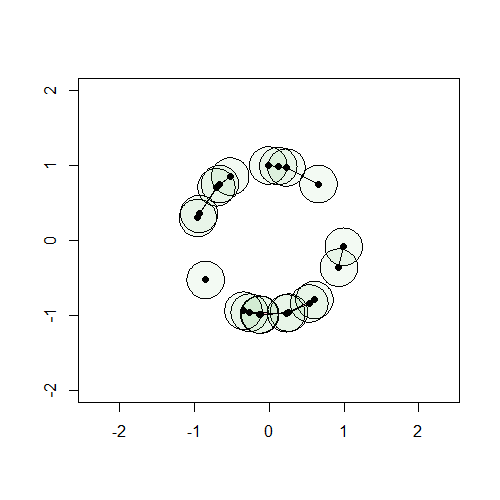}
         \includegraphics[scale=0.45]{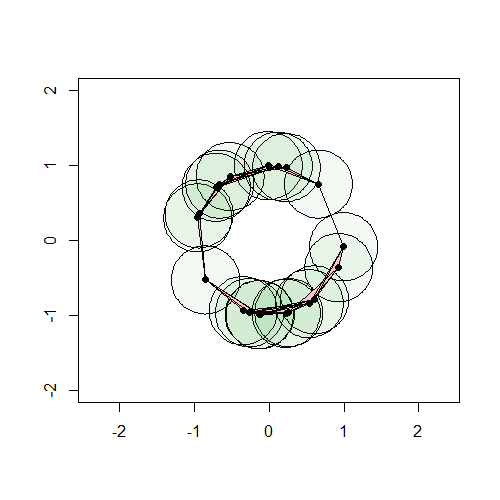}
         \includegraphics[scale=0.45]{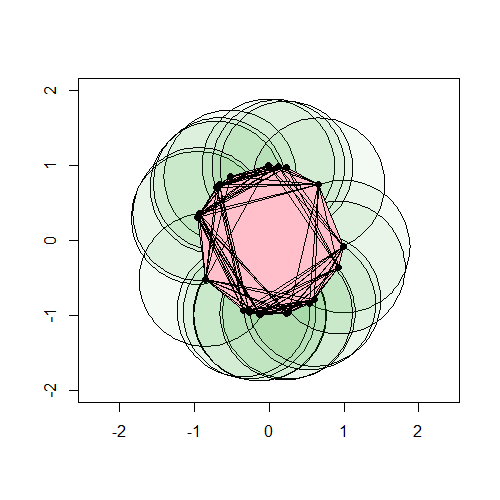}
        \caption{Example of Rips complex computation: (top-left) $\epsilon=0$; (top-right) $\epsilon=0.5$; (bottom-left) $\epsilon=0.9$; and (bottom-right) $\epsilon=1.8$.}
	\label{fig:loops}
\end{figure}

\begin{figure}[h!]
	\centering
	\includegraphics[scale=0.45]{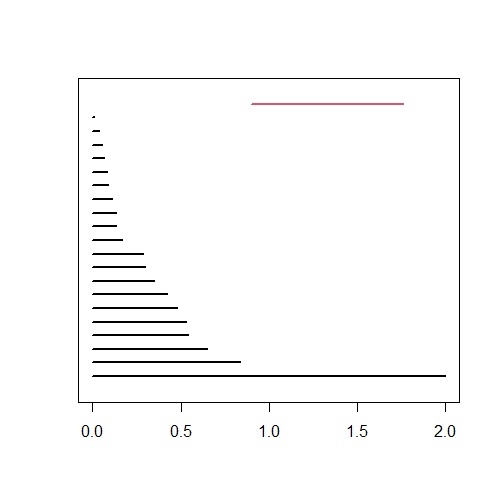}
	\caption{Persistence barcode: in black the $H_0$ features, and in red the $H_1$ feature. Filtration value (scale) is represented in the $x$-axis.}
	\label{fig:barcode}
\end{figure}

An example of persistent homology computation is given with the rips complex in Fig. \ref{fig:loops}, and the associated barcode in Fig. \ref{fig:barcode}. A loop is formed at $\epsilon=0.9$ (birth) and then covered at $\epsilon=1.8$ (death). It is represented by the red bar.

A more compact representation of the features persistence is the persistence diagram of $\mathbb{S}$, defined from
$$
\mathcal{PD}(\mathbb{S}) = \{(b_\gamma,d_\gamma) : \gamma \in H_k\},
$$
where $b_\gamma$ and $d_\gamma$ are the birth and death scales associated to the feature $\gamma$. In what follows, in the trajectories analysis,  we only consider one-dimensional features, i.e., $k=1$.

\begin{figure}[h!]
	\centering
	\includegraphics[width=0.5\linewidth]{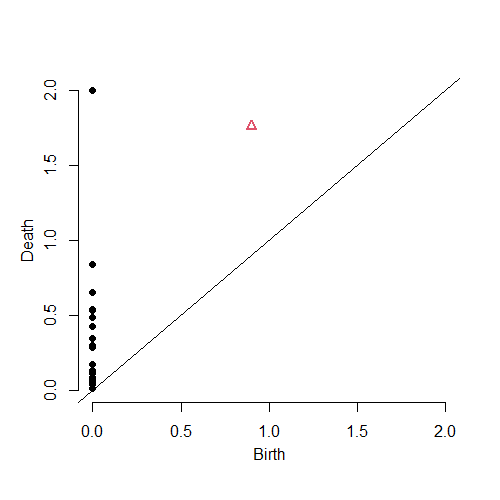}
	\caption{Persistence Diagram: in black the $H_0$ features, and in red the $H_1$ feature.}
	\label{fig:pd}
\end{figure}

The  persistence diagram associated with the Rips complex shown in Fig. \ref{fig:loops} is given in Fig. \ref{fig:pd}.
An equivalent representation of the persistence diagram consists in the so-called life-time diagram of $\mathbb{S}$, which is constructed by means of a bijective transformation $T(a,b) = (a,b-a),$ acting over
$\mathcal{PD}(\mathbb{S}),$ that is,
$$
\mathcal{LT}(\mathbb{S}):= \left\{(a,b-a) \in \mathbb{R}^2: (a,b) \in \mathcal{PD}(\mathbb{S}) \right\}.
$$
In order to use the persistence features in a machine learning approach, we construct the so-called {\em persistent image} of $\mathbb{S}$. First, observe that $\mathcal{LT}(\mathbb{S})$ is a finite set of $p$ points,
$$
\mathcal{LT}(\mathbb{S}) = \{(a_1,b_1-a_1),\ldots,(a_p,b_p-a_p)\},
$$
and such that $b_1 - a_1 \le b_2-a_2 \le \ldots \le b_p-a_p$. Then, consider a non-negative weighting function given by
\begin{align*}
     w:\mathcal{LT}(\mathbb{S}) &\rightarrow [0,1]\\
     (a_i,b_i-a_i) &\mapsto w(a_i,b_i-a_i) = \frac{b_i-a_i}{b_p-a_p}, \ \text{for} \ 1 \le i \le p.
\end{align*}

Finally, we fix $M$, a natural number, and take a bivariate normal distribution $g_{u}(x, y)$ centered at each point $u \in  \mathcal{LT}(\mathbb{S})$ with a variance $\sigma\mathbf{I}_2 = \frac{b_p - a_p}{M} \mathbf{I}_2 $ ($\mathbf{I}_2$ is the $2 \times 2$ identity matrix). A persistence kernel is then defined according to:
\begin{align*}
   \rho_{\mathbb{S}} : \mathbb{R}^2 &\rightarrow \mathbb{R} \\
   (x,y) &\mapsto \rho_{\mathbb{S}}(x,y) = \sum_{u \in \mathcal{LT}(\mathbb{S})} w(u)g_u(x,y).
\end{align*}

We associate  to a robot trajectory $\mathbb{S} \in \mathbb{R}^2$ a matrix in $\mathbb{R}^{M \times M}$ as follows:  let $\delta > 0$ be a non-negative, small enough real number, and then consider a squared region
$\Omega_{\mathbb{S},\delta}=[a,b] \times  [c,d]  \subset \mathbb{R}^2$, covering the support of $\rho_{\mathbb{S}}(x,y)$  up to a certain precision $\delta$, such that
$$
\iint_{\Omega_{\mathbb{S},\delta}} \rho_{\mathbb{S}}(x,y)\,dx\,dy \ge  1-\delta.
$$
Then, we consider two uniform partitions of the intervals
$$
a = p_0 \le p_1\le  \ldots \le p_M = b  \text{ and }
c = q_0 \le q_1\le  \ldots \le q_M = d.
$$
Finally, we express $\Omega_{\mathbb{S},\delta}$ from
$$
\Omega_{\mathbb{S},\delta} = \bigcup_{i=0}^{M-1} \bigcup_{j=0}^{M-1} [p_i,p_{i+1}] \times [q_j,q_{j+1}] = \bigcup_{i=0}^{M-1} \bigcup_{j=0}^{M-1} P_{ij}.
$$

The persistence image of $\mathbb{S}$ associated with the partition $\mathcal{P}=\{P_{ij}\}$ is then described by the $\mathbb{R}^{M \times M}$ matrix with elements:
$$
PI(\mathbb{S},M,\mathcal{P},\delta)_{ij} = \left(\iint_{P_{ij}} \rho_{\mathbb{S}}(x,y) dx dy\right) \text{ for } 0 \leq i,j \leq (M-1).
$$

\begin{figure}
    \centering
    \includegraphics[width=0.7\linewidth]{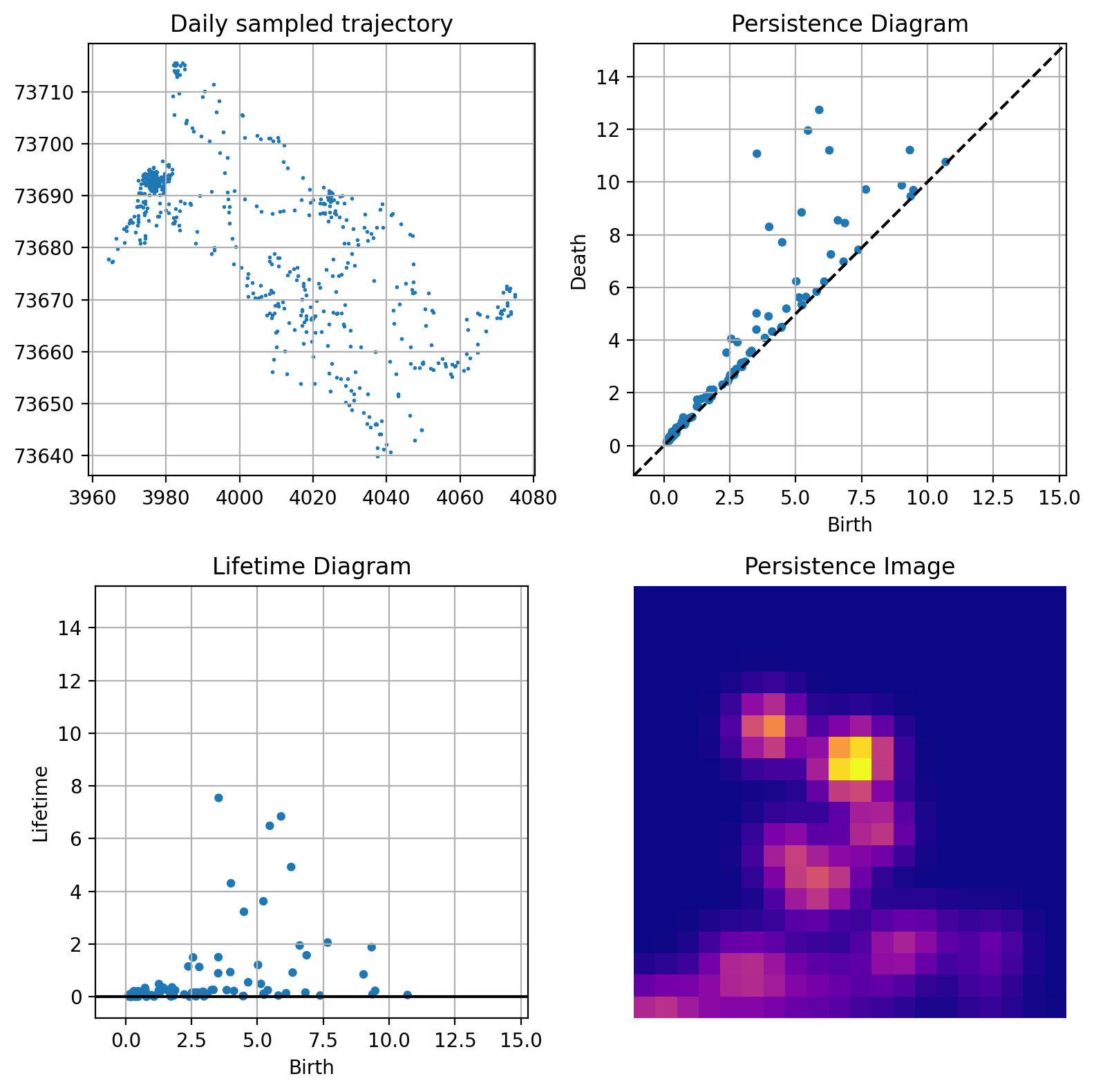}
    \caption{Topological analysis of a trajectory: (top-left) Trajectory; (top-right) Persistence diagram; (bottom-left) Lifetime diagram; and (bottom-right) Persistence Image.}
    \label{fig:pers}
\end{figure}

An example of persistence computation for a given trajectory is given in Fig.\ref{fig:pers}.

\subsection{Measuring persistence similarity}

Consider two data sets $\mathbb{S}_u$ and $\mathbb{S}_v$ representing two trajectories. A matching between two persistence diagrams, $\mathcal{PD}(\mathbb{S}_u)$ and $\mathcal{PD}(\mathbb{S}_v)$, is a map $\psi$, that reads:
$$
\psi: \mathcal{PD}(\mathbb{S}_u) \longrightarrow \ \mathcal{PD}(\mathbb{S}_v),
$$
such that $\forall \gamma = (b,d)\in \mathcal{PD}(\mathbb{S}_u)$,
\begin{align*}
    \psi(\gamma) &= (\psi_1(b),\psi_2(d)) \\
    &= (b',d') \in \mathcal{PD}(\mathbb{S}_v).
\end{align*}

The map $\psi$ associates each feature from $\mathcal{PD}(\mathbb{S}_u)$ to a feature from $\mathcal{PD}(\mathbb{S}_v)$.
The \emph{optimal matching} between $\mathcal{PD}(\mathbb{S}_u)$ and $\mathcal{PD}(\mathbb{S}_v)$ is a matching $\hat{\psi}$
$$
\hat{\psi}: \mathcal{PD}(\mathbb{S}_u) \longrightarrow \ \mathcal{PD}(\mathbb{S}_v),
$$
minimizing the transport cost $\mathcal{C}$ to move the features from $\mathcal{PD}(\mathbb{S}_u)$ to $\mathcal{PD}(\mathbb{S}_v)$:
\begin{align*}
    \mathcal{C}_{\mathtt{min}} =& \sum_{\gamma \in \mathcal{PD}(\mathbb{S}_u)}{\|\gamma-\hat{\psi}(\gamma)\|}_2 \\
    =& \sum_{(b,d) \in \mathcal{PD}(\mathbb{S}_u)} {\|\big(b-\hat{\psi}_1(b),d-\hat{\psi}_2(d)\big)\|}_2 \\
    =& \sum_{(b,d) \in \mathcal{PD}_k(\mathbb{S}_u)} \sqrt{\big(b-\hat{\psi}_1(b)\big)^2+\big(d-\hat{\psi}_2(d)\big)^2}.
\end{align*}

Then, to measure the degree of similarity between two trajectories $\mathbb{S}_u$ and $\mathbb{S}_v$ we consider the \emph{Wasserstein distance} \cite{R13,R20} between $\mathcal{PD}(\mathbb{S}_u)$ and $\mathcal{PD}(\mathbb{S}_v)$
$$
W\big(\mathcal{PD}(\mathbb{S}_u), \mathcal{PD}(\mathbb{S}_v)\big) = \sum_{(b,d) \in \mathcal{PD}(\mathbb{S}_u)} \sqrt{\big(b-\hat{\psi}_1(b)\big)^2+\big(d-\hat{\psi}_2(d)\big)^2},
$$
where $\hat{\psi}$ is the optimal matching between $\mathcal{PD}(\mathbb{S}_u)$ and $\mathcal{PD}(\mathbb{S}_v)$.

\begin{figure}
    \centering
    \includegraphics[scale=0.65]{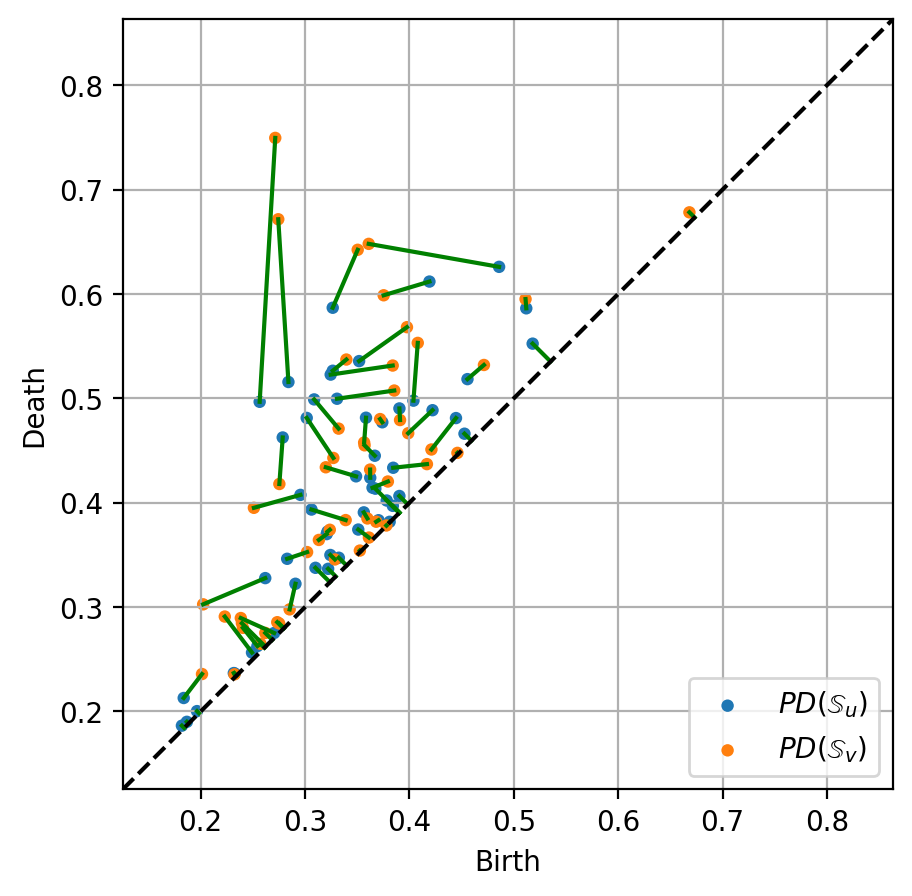}
    \caption{Optimal matching between two persistence diagrams related to two robot trajectories}
    \label{fig:match}
\end{figure}

An example of matching between the persistence diagrams of two trajectories is given in Fig. \ref{fig:match}.

\subsection{Barycenters of persistence diagrams}

Consider now a collection $\mathbb{S}_1 \dots \mathbb{S}_n$ of trajectories with their associated diagrams $\mathcal{PD}_1 \dots \mathcal{PD}_n$.\\
Since the space of persistence diagrams equipped with the Wasserstein distance, the \emph{ Wasserstein space}, is not a linear space, the notion of  barycenters \cite{R21} can be extended for the persistence diagrams using the so-called \emph{Frechet mean} \cite{R22}, which always exists in the context of averaging finitely many diagrams.

The Frechet mean of $\mathcal{PD}_1 \dots \mathcal{PD}_n$ is any diagram minimizing the map
$$
\mathcal{E} : \mu \mapsto \sum_{i=1}^n W(\mu, \mathcal{PD}_i)^2.
$$

The computation of the barycenter $\mu$ has proven to be challenging, and multiple approaches can be used, such as the Sinkhorn algorithm \cite{R23}. We will use the one  based on the Hungarian algorithm presented in \cite{R22} and consider Partial Optimal Matchings \cite{R24}, as the diagrams may not be of the same size. In this case, points from the diagonal are matched with the remaining (exceeding) points.

In our case, we estimate the barycenters of a finite family of persistence diagrams, taking a Lagrangian approach by tracking the individual points of the diagrams. Given a collection $\mathcal{PD}_1 \dots \mathcal{PD}_n$ of persistence diagrams, we proceed as follows:
\begin{enumerate}
	\item Intialize the estimation $\mu$ of the barycenter at a certain diagram $\mu = \mathcal{PD}_{i_0}$.
	\item Compute the optimal partial matchings $\psi_1 \dots \psi_n$, between $\mu$ and $\mathcal{PD}_1 \dots \mathcal{PD}_n$ respectively.
	\item Compute the updated barycenter $\hat{\mu}$, by averaging the transport of each point in the barycenter $\mu$
$$
\hat{\mu} = \{y = \frac{1}{n} \sum_{i=1}^n \psi_i(x), x\in \mu \}.
$$
	\item If $\hat{\mu}$ minimizes $\mathcal{E}$, return $\hat{\mu}$. Otherwise, update $\mu = \hat{\mu}$ and go back to 2.
\end{enumerate}

\begin{figure}
    \centering
    \includegraphics[scale=0.5]{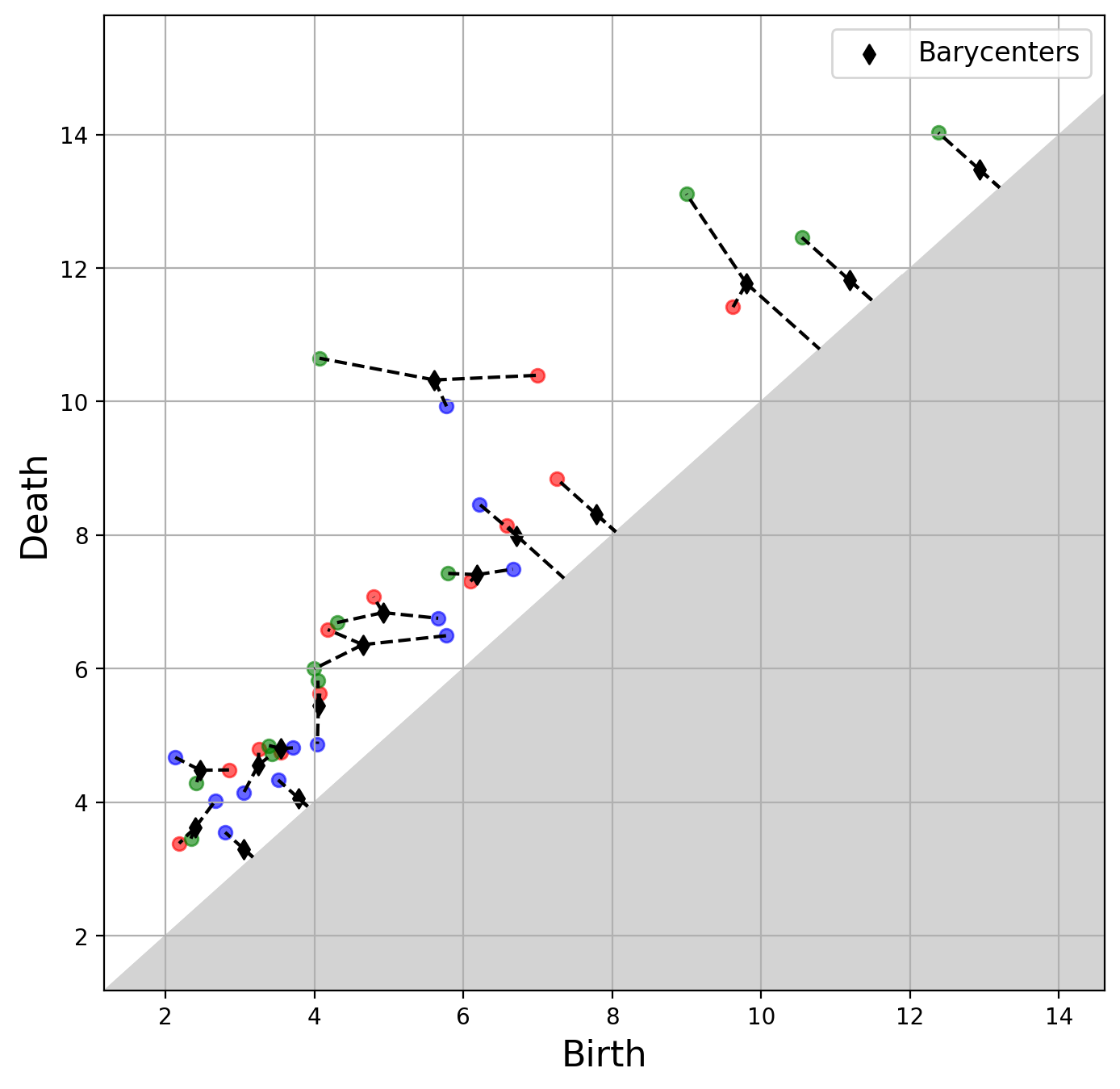}
    \caption{Barycenter (in black) of three persistence diagrams (red, blue and green)}
    \label{fig:bar}
\end{figure}

An example of  a  barycenter of three persistence diagrams is given in Fig. \ref{fig:bar}.

\subsection{Classification}

Image classification is a procedure that is used to automatically categorize images into classes by assigning to each image a label representative of its class. A supervised classification algorithm requires a training sample for each class, that is, a collection of data points whose class of interest is known. Labels are assigned to each class of interest. The classification problem applied to a new observation (data) is thus based on how close a new point is to each training sample. The Euclidean distance is the most common metrics used in low-dimensional datasets. The training samples are representative of the known classes of interest to the analyst. In order to classify the persistence images, we considered the logistic regression algorithm.

Consider a training set $\big(\mathcal{X}_i\big)_{i=1}^{n}$ of flattened persistence images, i.e., $M \times M$-component vectors, computed from a set $\big(\mathbb{S}_i\big)_{i=1}^{n}$ of trajectories as described earlier. Associated is a list $(\mathcal{Y}_i)_{i=1}^{n}$ of binary labels $\{0,1\}$, describing whether an image $\mathcal{X}_i$ is in the interest set or not.

The training of the \emph{$\mathcal{L}_2$-penalized logistic regression binary classifier} is then the minimization of a cost function as described in the following optimization problem:
$$
\min_{\omega,c} \frac{1}{2} \omega^T \omega + C \sum_{i=0}^p \log\bigg(\exp\Big(\mathcal{Y}_i\big(\mathcal{X}_i^T \omega + c \big)\Big)+1\bigg).
$$

Here $\omega$ are the weights we optimize over, $c$ a Bernouilli mean vector of the weights, and $C$ an inverse regularization parameter.
Once trained, the model is evaluated on a unseen set of flattened persistence images.
The metrics used for the model evaluation is the \emph{Accuracy Score} defined as the number of correct predictions over the number of samples.

\section{Results}

\subsection{Determination of the patch in which the robot is located}

We first want to predict whether a robot is in a certain patch. For that purpose we choose one parcel as a target, and train a classification model as described in Section 2.6. The complete dataset consists of daily trajectories for $240$ days. For each day a persistence image is computed, that will be used as input for the model (a sample is depicted in Figure \ref{fig:pers}).

The samples are labelled according to the considered patch, $1$ if the robot is in the target patch, and $0$ otherwise. The dataset is split into $65\%$ for training and $35\%$ for testing.
The proposed classifier achieves an $80\%$ accuracy score in predicting the patch at which the robot is, based on the persistence images.

\subsection{Maintenance prediction}

Then, we consider daily trajectories in the same patch, consisting of $50$ samples. For each day, a persistence image is computed, that will be used as input in the classifier.
The periods considered here are the ones in between two consecutive maintenance operations of the robot. The samples are labelled $0$ if they are associated to a day before the maintenance date, $1$ otherwise. The dataset is split into $65\%$ for training and $35\%$ for testing. The model achieves a $90\%$ accuracy score predicting the period associated to the the sampled trajectories, proving that robot trajectories exhibit a topological pattern when maintenance applies, fact that could be used for predictive maintenance purposes. 

\begin{figure}
    \centering
    \includegraphics[width=\linewidth]{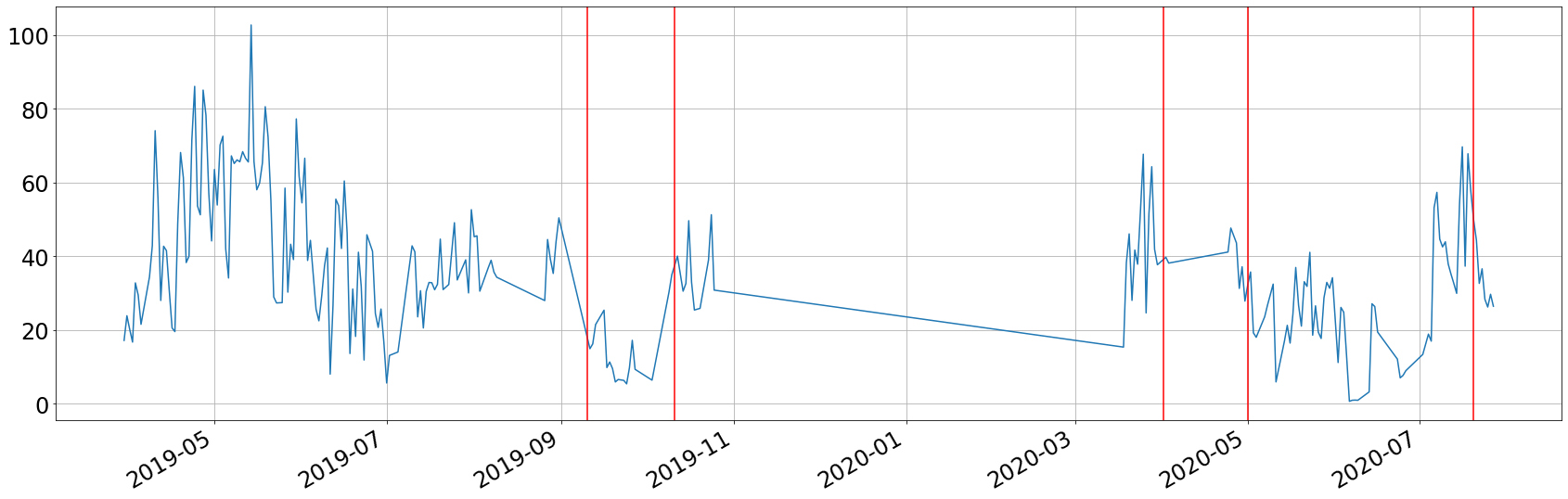}
    \caption{Time series of the Wasserstein distance between the persistence diagrams for consecutive daily trajectories: in red the maintenance events. }
    \label{fig:wass}
\end{figure}

\begin{figure}
    \centering
    \includegraphics[width=\linewidth]{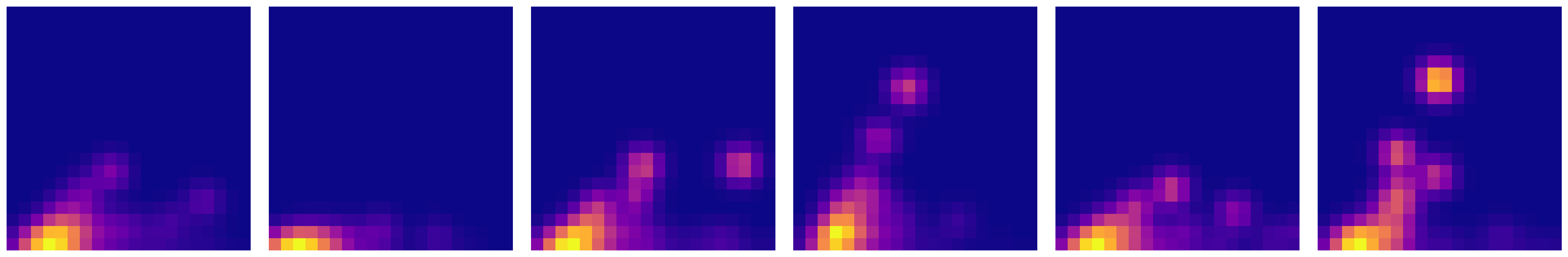}
    \caption{Persistence images of the barycenters computed for each period}
    \label{fig:second}
\end{figure}

Figure \ref{fig:wass} depicts the Wasserstein distance between the persistence diagrams for consecutive daily trajectories, with the maintenance operation emphasized in red, whereas Fig. \ref{fig:second} shows the barycenters of each period between consecutive maintenance operations. As it can be noticed from the persistence images in Fig. \ref{fig:second}, maintenance operations affect the topology of the trajectory, as it was expected from the fact that classification performs successfully as just reported. 

\begin{figure}
    \centering
    \includegraphics[scale=0.3]{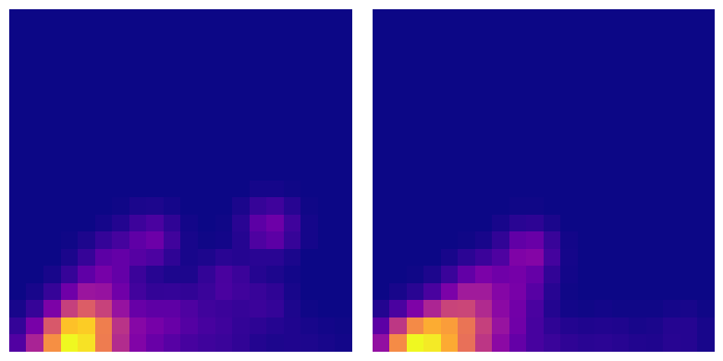}
    \caption{Persistence images of the two half-intervals related to the first period whose persistence image was the first image in Fig. \ref{fig:second}}
    \label{fig:compara}
\end{figure}

To better support our hypothesis about the effect of maintenance on the trajectory topology, we consider the first operation interval, the one before the first maintenance, that correspond to the first persistence image in Fig. \ref{fig:second} (left), and divide it in two parts with identical length. Then, the associated barycenters in both half intervals are obtained. Both are represented in Fig. \ref{fig:compara}. As it can be noticed, both of them resemble very much to the one associated to the whole interval (the first picture in Fig. \ref{fig:second}), all them (both in Fig. \ref{fig:compara} and the first in Fig. \ref{fig:second}) are significantly different to the second image in Fig. \ref{fig:second} that represents the trajectory topology after the first maintenance operation. These results support again our assumption on the effect of maintenance on the trajectory topology.

\section{Conclusions}

The characterization of the trajectories followed by the robot based on the geographical location proves to be a reliable method to differentiate between different environments affecting the robot motion. Then, over a single patch, the classification was proved being efficient to detect the changes in the robot signature related to maintenance events.

The proposed topology-based framework for sampled trajectories seems a very pertinent, powerful and intrinsic way of quantifying, characterizing and analyzing the topological and geometrical nature of the robot's pathways. The strength of the framework relies on both the topology description of the trajectory at multiple scales, and the use of metrics features that can be combined with machine learning.


\end{document}